\documentclass{ifacconf}
\usepackage{amsmath}

\usepackage{amssymb}
\usepackage{algorithm}
\usepackage{algorithmic}

\newtheorem{definition}{Definition}[section]

\newtheorem{lemma}{Lemma}[section]

\usepackage{graphicx}

\usepackage{xcolor,soul}
\newcommand{\colnode}{\ensuremath{\{\xi ^1, \xi ^2, \dots, \xi ^N\} }}
\usepackage{savesym}
\savesymbol{AND}
\usepackage{amssymb,amsmath}
\usepackage{graphicx}      % include this line if your document contains figures
\usepackage{natbib}        % required for bibliography
\begin{document}
\begin{frontmatter}

% new title
\title{
	A New Distribution-Free Concept for
	Representing, Comparing, and Propagating Uncertainty in Dynamical Systems with Kernel Probabilistic Programming\thanksref{footnoteinfo}} 

% \title{A \st{New Distribution-Free Concept} Kernel Mean Embedding Approach \st{for} to
% 	Representing, Comparing, and Propagating Uncertainty in Dynamical Systems \st{with Kernel Mean Embedding}\thanksref{footnoteinfo}} 
% % Title, preferably not more than 10 words.

\thanks[footnoteinfo]{This project has received funding from the European Union’s Horizon 2020 research and innovation programme under the Marie Skłodowska-Curie grant agreement No 798321, the German Federal Ministry for Economic Affairs and Energy (BMWi) via eco4wind (0324125B) and DyConPV (0324166B), and by DFG via Research Unit FOR 2401.
% \jz{everybody, please add your support acknowledgement}
}

\author[First]{Jia-Jie Zhu} 
\author[First]{Krikamol Muandet}
\author[Second]{Moritz Diehl}
\author[First]{Bernhard Sch\"olkopf}

\address[First]{Empirical Inference Department,\\Max Planck Institute for Intelligent Systems, T\"ubingen, Germany.\\(e-mail: \{jzhu, krikamol, bs\}@tuebingen.mpg.de)}
\address[Second]{Department of Microsystems Engineering,\\ 
   University of Freiburg, Freiburg, Germany.\\(e-mail: moritz.diehl@imtek.uni-freiburg.de)}

\begin{abstract}            % Abstract of not more than 250 words.
This work presents the concept of kernel mean embedding and kernel probabilistic programming in the context of stochastic systems. 
We propose formulations to represent, compare, and propagate uncertainties for fairly general stochastic dynamics in a distribution-free manner. 
The new tools enjoy sound theory rooted in functional analysis and wide applicability as demonstrated in distinct numerical examples.
The implication of this new concept is a new mode of thinking about the statistical nature of uncertainty in dynamical systems.
\end{abstract}
\end{frontmatter}
%===============================================================================

\section{Introduction}\label{sec:intro}
Classic stochastic control methods such as LQG hinge on the mathematical fact that the family of Gaussian distributions is closed under an affine transformation. %the affine transformation of a Gaussian distribution is, again, Gaussian.
This allows uncertainty to be propagated in a tractable manner under the Gaussianity assumption.
Robust control methods, such as the classic tube  model predictive control, also rely on the linearity of dynamics to propagate the polytopic uncertainty.
However, when we move beyond those assumptions to the territories of nonlinear dynamics and non-Gaussian noise, uncertainty propagation becomes difficult or even intractable.

A central component in robust and stochastic control is how to \textit{represent} the system uncertainty. To this end, point estimate, ellipsoidal uncertainty set, Gaussian distribution, or randomized computer simulations have been used in various applications. 
Along with those, a few families of \textit{uncertainty propagation} methods have been proposed, e.g., generalized polynomial chaos approximation, Gaussian processes.
The representation and propagation affect control as well as system identification. For example, a parameter estimation problem typically uses the likelihood function as its criterion for \textit{goodness-of-fit}, which requires assuming the distribution family.

\begin{figure}
	\centering
	\includegraphics[width=0.9\columnwidth]{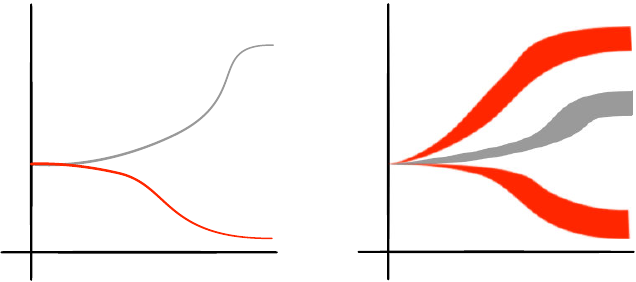}
	\caption{An illustrative example of deterministic (left) and stochastic (right) dynamical systems. Notice the system in (right, red) is stochastic and bi-modal.}
	\label{fig:intro}
\end{figure}

Consider an illustrative sketch in Figure \ref{fig:intro}. 
How can we measure the differences between system models?
If the dynamical system of interest is
deterministic (left), then we can simply compare the solutions of the systems in the sense of Euclidean distance.
%  between the states of the two systems. 
Now we consider Figure \ref{fig:intro} (right), which illustrates the evolution of two \textit{stochastic} dynamical systems (with arbitrary distribution). Due to the stochasticity, the
solutions are \textit{distributions}. Comparison in Euclidean distance is no longer feasible.
How do we incorporate statistical information without imposing strong distributional assumptions? 
The question we address is how to \textit{represent, compare, and propagate} the randomness in dynamical systems in a quantitative way.
% \footnote{This is also applicable if we instead use set-theoretical
% calculus rather than probability theory.}
% In full generality, stochastic processes in systems may be time-varying, non-stationary, and, of unknown distribution. 

This work proposes to embed uncertain state
distributions into the reproducing kernel Hilbert spaces (RKHS), enabling  quantification and algebraic operations therein.
The main contributions of this work are the following.
\begin{itemize}
	\item We introduce the RKHS embedding method of kernel probabilistic programming in the context of uncertain dynamical systems. The resulting formulations of representation, comparison, and propagation methods are our main contributions.
	\item
	Through distinct numerical examples, we demonstrate the flexible and distribution-free nature of our method as a unifying tool for dynamical systems.
	\item We propose a recursive reduced set method to propagate the system uncertainty by iteratively solving an optimization problem. This forms fixed-size kernel expansions rather than allowing the number of uncertainty realizations to explode over time.
	% , as is commonly encountered in robust control [e.g., \cite{scokaert1998min}; \cite{tempo2012randomized} Chapter~12; \cite{campi2019scenario}].
\end{itemize}
\subsubsection{Notation:}
In this work, the state of uncertain dynamical systems are denoted by $x(\tau,\xi)$, where $\tau$ is time and $\xi$ is the uncertainty in the system, e.g., a disturbance process. The Stieltjes integral $\int \cdot \ d P$ denotes the integration w.r.t. either a continuous or a discrete probability measure $P$. 
Symbol $\mathcal{X}$ often denotes a nonempty set and $\mathcal{H}$ a reproducing kernel Hilbert space (RKHS). We write $\xi\sim P$ to denote that the random variable (RV) $\xi$ follows the distribution law $P$.
% of real-valued functions on $\mathcal{X}$, endowed with a reproducing kernel $k:\mathcal{X}\times\mathcal{X}\to\mathbb{R}$ and an inner product $\langle \cdot,\cdot \rangle_{\mathcal{H}}$.

\section{Background \& related work}\label{sec:background}

\subsection{Uncertainty quantification in dynamical systems}
\label{sec:uq}
Stochastic dynamical system refers to a system whose evolution is affected by non-deterministic disturbances. 
% Mathematically, this may be modeled as differential equations with uncertain components. 
To illustrate our idea concretely, let us consider a simple example of stochastic ODE that was studied in \cite{Xiu2003}:
\begin{equation}
% \begin{aligned}
\dot x(t)  = \xi\ x,\quad x(0) = x_0,
% \end{aligned}
\label{eq:motivEx}
\end{equation}
where the parameter $\xi$ is uncertain. 
This uncertainty may enter as (a) time-invariant, or, more generally (b) time-varying stochastic process.
For a moment, let us consider case (a) and the parameter follows a certain distribution law  $\xi\sim P_\xi$. This is hence an initial value problem (IVP) with uncertain parameter.
The problem of quantifying the distribution of the solution $x(t,\xi)$ to the IVP~\eqref{eq:motivEx} is typically referred to as the \textit{uncertainty quantification} in dynamical systems.

One somewhat trivial approach of quantifying the solution is the Monte Carlo sampling that samples \colnode\ from the underlying distribution. Then we deterministically integrate the ODE with those realizations of the uncertain parameter to obtain solutions $\{x(t,\xi^i)\}_{i=1}^N,\ \forall t$. We may later extract the moment statistics of $x(t,\xi)$ by its Monte Carlo estimation.

In numerical analysis, one representative thread of works in uncertainty quantification is the generalized polynomial chaos (gPC) method.
It expands the solution of IVP \eqref{eq:motivEx} as a series in orthogonal polynomial basis functions $\{\phi_i(\xi)\}_{i=1}^N$,
$$
x(t,\xi) = \sum_{i=1}^\infty \alpha_i \phi_i(\xi).
$$
This expansion is mathematically elegant in that the basis $\phi_i(\xi)$'s
% , which are random functions of random variable (RV) $\xi$, 
account for the stochasticity and the expansion coefficients $\alpha_i$'s are deterministic.
From this point onward, we can either follow Galerkin's method propagate the coefficients $\alpha_i$ through the dynamical systems as in \cite{Xiu2003}, or, use the so-called stochastic collocation to numerically integrate the IVP \eqref{eq:motivEx} at certain collocation nodes \colnode\ as in \cite{Xiu2009}.
% We shall see that our approach generalizes the orthogonal polynomial basis via the tool of kernel mean embeddings.

Another well-developed methodology in Bayesian statistics is the Gaussian process (GP). Intuitively, GP generalizes the Gaussian distribution to the distribution of functions. 
% One way to apply it to model system uncertainty is to represent the state evolution equation by a GP prior. 
For example, dynamics described by a difference equation can be modeled by a GP prior, i.e., $x_{t+1} - x_t=f(x_t) \sim \mathcal{GP}(\mu(x_t), \sigma(x_t))$ where $x_t$ is a shorthand for $x(t,\xi)$. Given data samples, it can be shown that the posterior predictive distribution for an unseen state is also Gaussian. This allows us to quantify the distribution of solution $x(t,\xi)$.
%  $P\left(f(x^*)  | x^*, \{x^i,f(x^i)\}_{i=1}^N \right)$ given the observed data $\{(x^i,f(x^i))\}_{i=1}^N$.
% \begin{eqnarray}
% 	\begin{array}{ll}
% 		P&\left(f(x^*)  | x^*, \{x^i,f(x^i)\}_{i=1}^N \right) \\
% 		& =\mathcal{N}\left(\mu(x^*), \sigma (x^*)\right).
% 	\end{array}
% \end{eqnarray}
% where $\mu(x^*), \sigma (x^*)$ are the mean and covariance functions of the GP. 
We refer to \cite{rasmussen2006gp} for an accessible introduction.

% neural net dynamics, keep it short
% Yet another increasingly popular modeling choice is to use (deep) neural networks (NN) to represent dynamical systems. Mathematically, this is a parametric system model $f _ \theta: x_t \to x_{t+1}$ where the parameters $\theta$ are the weights of the NN.
% Although a NN dynamics model is inherently deterministic, there are efforts to incorporate uncertainty descriptions such as \cite{gal2016improving}.

Mathematically speaking, gPC and GP are both surrogate function classes enabled by function approximation theory.
This paper does not focus on analyzing the function approximation aspect.
Instead, the embedding method we shall propose in Section \ref{sec:approach} calls for a shift in our ways of thinking about statistical distributions. 
% Intuitively, we represent the (stochastic) dynamical system solution $x(t,\xi)$ as a deterministic member of a Hilbert space thus allowing rich operations. We    

\subsection{Goodness-of-fit measure for system identification}
System identification studies how to construct mathematical models of dynamical systems from observed data.
% Depending on the model in question, 
% Identification methods may involve methodologies such as parameter estimation, variability estimation, optimal experimental design.
While this paper does not directly propose a system identification algorithm, we show how the proposed concept can impact how we analyze and compare the models in terms of \textit{goodness-of-fit}.
To give readers a concrete example of this new concept, we revisit the parameter and variability estimation (PVE) proposed by \cite{Mohammadi2015} in Section~\ref{sec:exp2}.
This may be thought of as an alternative to the least square (LSQ) estimation, whose underlying assumption is that the system is following an additive Gaussian distribution with fixed variance,
$y = f(x) + \xi, \ \xi \sim N (\bar \xi_\mathrm{LSQ}, \sigma_\mathrm{LSQ}).$
In contrast, PVE is based on the \textit{robust optimization} idea that the disturbance should lie within an ellipsoidal uncertainty set
$\xi \in \mathcal E (\bar \xi _\mathrm{PVE} , Z _\mathrm{PVE})$
but not making any assumptions on the distribution family. The PVE method then formulates the identification of the uncertain ellipsoid as a semidefinite programming (SDP) problem. More details are provided in that paper.

It can be relatively difficult to compare an LSQ point estimate (or MLE in general) with e.g. PVE because they do not share the same likelihood .
This paper proposes such a unifying framework of comparing different system models' goodness-of-fit.
% In this work, we answer the question of how to qualitatively compare different system models such as LSQ and PVE in an unified framework.

\subsection{Reproducing kernel Hilbert space (RKHS) embedding}\label{sec:bg_kernel}
% \jz{insert kernel is a bi-variate function, blahblah}

% The most essential tool in representing uncertainty for our framework is the kernel mean embedding (KME) of probability distributions [\cite{Smola07Hilbert,Muandet2017}] defined in terms of positive definite kernel functions [\cite{scholkopf2002learning}].
A kernel is a real-valued bivariate function $k(\cdot,\cdot):\mathcal{X}\times\mathcal{X}\to\mathbb{R}$. It is said to be positive definite if $\sum_{i,j=1}^n\alpha_i\alpha_jk(x_i,x_j) \geq 0$ for any $n\in\mathbb{N}$, $(\alpha_1,\ldots,\alpha_n)\in\mathbb{R}^n$, and $(x_1,\ldots,x_n)\in\mathcal{X}^n$. 
% It can be shown that $k$ is positive definite if and only if it is  a reproducing kernel [\cite{scholkopf2002learning}].
% \jz{need to dumb down here}
In addition, it is a \textit{reproducing kernel} of an RKHS $\mathcal{H}$ if (i) $\forall x\in\mathcal{X}$, $k(x,\cdot)\in\mathcal{H}$ and (ii) $\forall x\in\mathcal{X},f\in\mathcal{H}$, $f(x) = \langle f,k(x,\cdot)\rangle_{\mathcal{H}}$.
The latter is known as the \emph{reproducing property} of $\mathcal{H}$.
Choosing $f = k(x',\cdot)$ for some $x'\in\mathcal{X}$ and applying the reproducing property yield the \textit{kernel trick}
\begin{equation}
 k(x, x') = \langle k(x,\cdot), k(x',\cdot)\rangle_{\mathcal{H}} = \langle \phi(x),\phi(x')\rangle_{\mathcal{H}}.
 \label{eq:trick}
\end{equation}
That is, we can view the kernel evaluation $k(x,x')$ as a generalized similarity measure between $x$ and $x'$ after mapping them into the feature space $\mathcal{H}$.
We refer to \(\phi\) defined above as a \emph{canonical feature map} associated with the kernel $k$.
One of the most common kernels on $\mathbb{R}^d$ is the Gaussian kernel
\begin{equation}
	k(x,x') = \exp\left(-\frac{1}{2\sigma^2}\|x-x'\|_2^2\right), \quad x,x'\in\mathbb{R}^d,
\end{equation}
where $\sigma > 0$ is a bandwidth parameter.
% This similarity measure preserves statistical information of data.

An important application of kernel methods is in representing probability measures via \emph{kernel mean embedding} (KME) [\cite{Smola07Hilbert}]. 
This line of work can be thought of as a systematic way of endowing unstructured data with representations in a Hilbert space to provide the ability to perform algebraic operations therein.
Mathematically, we define the KME of a random variable $X$ as follows.
\begin{definition}{(Kernel mean embedding)}
Given random variable $X\sim P$ and  kernel \(k\) for which $\mathbb{E}_X\sqrt{k(X,X)} < \infty$, we
define the kernel mean embedding of $X$ as a function
\[
\mu^{k}_X(\cdot) = \int  k(x, \cdot) \ dP(x).
\] This function is a member of the RKHS,
\(\mu^{k}_X \in \mathcal{H}\) associated with the kernel $k$.
\end{definition}
In the rest of the paper, we follow the convention in kernel machine learning to simply write the function $\mu^{k}_X (\cdot)$ as $\mu_X$ to emphasize that it is an element of the RKHS $\mathcal{H}$. 
It has been shown that if $k$ belongs to a class of kernels known as \emph{characteristic kernels}, then $\mu_X$ uniquely determines the distribution $P$; see, e.g., \cite{Fukumizu04:DRS,Sriperumbudur10:Metrics}.

To help readers get a concrete understanding, we outline common kernels and the information their KME preserve in Table \ref{table:kernel}. We then give examples of the explicit forms of the KMEs.
\begin{table}[hb]
\begin{center}
\caption{Common kernels and what statistical information their KME preserve. More examples can be found in \cite{Muandet2017}}\label{table:kernel}
  \resizebox{\columnwidth}{!}{
\begin{tabular}{lll}
\hline
Linear & $k(x, x') = x^\top x'$ & Mean of distribution\\
Polynomial & $k(x, x') = (x^\top x' +1) ^p $ & Up to \(p\)-th moments \\
Gaussian & $k(x, x') = \exp\left(-\frac{\|x-x'\|_2^2}{2\sigma^2}\right)$ & All information\\
Exponential & $k(x, x') = \exp (x^\top x') $ & All information\\\hline
\end{tabular}}
\end{center}
\end{table}

\subsubsection{Example (Second-order polynomial kernel embedding)}
Suppose the kernel function in question is the polynomial kernel of order two $k(x, x') = (x^\top x'  + 1)^2$, the KME is given by 
\begin{equation} \label{eq:p2_kernel}
\begin{array}{ll}
\mu_{X}  & = \int k(x , \cdot ) d P(x) = \int \left(x^\top (\cdot) + 1\right)^2\ d P(x)\\
& = (\cdot) ^\top \mathbb{E}{x x^\top}  (\cdot) + 2 \mathbb{E} {x}^\top(\cdot) +1.
\end{array}
\end{equation}
This shows that the  RKHS associated with this KME consists of quadratic functions whose coefficients preserve statistical information up to the second order (mean and variance), but not higher. In general, $p$-th order polynomial kernel embeddings preserve information up to the $p$-th order. A richer kernel embedding, e.g. Gaussian kernel embedding, may preserve information up to infinite order.

\emph{Remark}
% (a).
% In stochastic control approaches, \textit{nominal} state and variance are typically considered.
% KME provides a generalized way to consider higher order information.
% (b).
Readers familiar with polynomial approximation may recognize that the integrand in \eqref{eq:p2_kernel} can be expanded in certain Wiener-Askey polynomial bases.
However, 
% as we move beyond polynomial kernels to rich kernels such as Gaussian kernel, 
our method differs from the philosophy of gPC expansion in that it does not seek to use finite-order truncation for approximating functions. Rather, we make use of the kernel trick~\eqref{eq:trick} to represent similarity in data even in infinite-dimensional feature space. This gives rise to the power of \textit{kernel machine learning}.

% e KME will henceforth be represented by its feature map by virtue of \eqref{eq:eigen_exp}.
% \begin{equation}
% \mu_x = \int   \phi(x)\ d P_x,
% \end{equation}
% where $P_x$ is the law of distribution of $X$ and the integral is a Stieltjes integral.
\subsubsection{Example (Exponential kernel embedding)} 
Given the exponential kernel $k(x, x') = e^{\langle x, x'\rangle} $, the KME is 
  \[
  \mu_X = \int e^{\langle x, \cdot\rangle} d P(x).
  \] 
  This is the moment-generating function. Notably, replacing $x$ by $-x$ yields the Laplace transform.

% kme estimator
The following result in \cite{Song08:Thesis} gives the consistency of a sample-based estimator $\hat \mu_{X}$ for $\mu_{X}$.
\begin{lemma}{(\cite{Smola07Hilbert}; \cite{Song08:Thesis}; estimator for KME)}
Let us denote by $X$ a random variable and 
$\{x^i\}_{i=1}^N$ its i.i.d samples.
Then,
\begin{equation}
\hat \mu _ {X} := \frac1N \sum_{i=1}^N k(x^i, \cdot) \to \mu _ { X}
\end{equation}
as $N \to \infty$ with probability $1$.
% Furthermore, the convergence rate is $\mathcal O (1/\sqrt{N})$ and is independent of dimensions.
\end{lemma}

%\jz{@krik: please check this prop. especially the condition on f and samples having to be iid?}

Furthermore, \cite{Scholkopf2015,SimonGabriel2016} proved the estimation consistency for KME of transformations of RVs in more general conditions. They term their approach \textit{kernel probabilistic programming} (KPP). 
\begin{lemma}{(KME consistency)}\label{thm:kme_consis}
Suppose $f: \mathcal X \to \mathcal Z$ is a continuous function, $k_x, k_z$ are continuous kernels on $\mathcal X$ and $\mathcal Z$. Under mild conditions 
[cf. Theorem 1 of \cite{SimonGabriel2016}]
, the following is true.
$$
\text{If }\hat \mu ^{k_x} _ {X} \to \mu ^{k_x} _ {X},\text{ then } \hat \mu ^{k_z} _ {f(X)} \to \mu ^{k_z} _ {f(X)}.
$$
%\jz{MD: explain how KPP is different from KME @krik, could you help with this?}
\end{lemma}
Notably, we do not require samples to be i.i.d.
This result equips us with algebraic tools to learn an embedding of $f(X)$ directly from that of $X$. 
% This applies to functions of multiple RVs, i.e., $f(X_1,\ldots,X_m)$ for $m>1$, as well. 
In the rest of the paper, by KPP we mean \emph{the RKHS embedding method that performs algebraic operations on KMEs via transformations of random samples}.

% ### existing approach on conditional embeddings
% {\color{blue}
% Before presenting our approach, we point out that KME has previously been applied to learning dynamics in machine learning.
In the context of reinforcement learning, e.g., in \cite{Nishiyama12:POMDPs,Grunewalder12:MDPs,boots2013hilbert}, RKHS embeddings of \emph{conditional} distributions, which is different from ours, were used to learn dynamics. 
We share the common thread of using RKHS embeddings while differ in a few important aspects, e.g., our use of KPP, numerical integration for continuous-time systems.
See \cite{Song2013} for more details and references.
% }

\section{Approach}\label{sec:approach}
\subsection{Representing uncertainty with RKHS embeddings}
KPP introduced in the last section gives us a powerful tool to represent distributions without any parametric assumption.
In the context of dynamical systems, if we view the evolution of the system uncertainty as transformations of random variables (RV), then KPP naturally becomes a tool to propagate system uncertainty.
An important motivation of our methodology is that it shall be \textit{distribution-free}, i.e., it shall not impose assumptions on the uncertainty distributions (e.g., Gaussianity).
% This allows us to represent rich information the uncertainty distribution may be highly skewed.

In a nutshell, we represent the distribution of $x(\tau,\xi)$, the state of the dynamical systems (continuous or discrete time), by its KME and the corresponding sample-based estimator given by
\begin{equation}
\begin{array}{cc}
\mu_{x(\tau,\xi)}  = \int k( {x(\tau,\xi)}, \cdot ) d P (\xi),\\
\hat \mu _ {x(\tau,\xi)}   = \sum_{i=1}^N \alpha_i k({x(\tau,\xi^i)}, \cdot),
\end{array}
\label{eq:kme}
\end{equation}
where a simple choice is $\alpha_i=\frac1N$.
As discussed in Section \ref{sec:bg_kernel}, KME with second-order polynomial kernel preserves  (nominal state) and second (variance) order information commonly used in stochastic control.
In this light, we may view our method as a generalization of Monte Carlo moment estimation.
% Yet it can generalize to capturing higher order information such as skewness (3rd), kurtosis (4th), etc, up to infinite moments when used with characteristic kernels.
% \vspace{-0.1cm}
\subsection{Goodness-of-fit measure for uncertain system models}
% motivate this sec
As suggested in Figure~\ref{fig:intro} (right), it may be difficult to quantitatively compare stochastic system models directly.
% How do we quantitatively compare their statistical properties?~\footnote{We may certainly compare higher order moments by Monte Carlo estimation. But as we show in this paper, the KME method generalizes Monte Carlo sampling.}
% This distance measure of two uncertain systems can be applied to, for example, evaluating the results of system identification procedures. 
For example, say we have identified an LSQ point estimate $\hat \xi_1$ and another estimation described by a distribution in uncertain parameter $\hat \xi_2 \sim P_2$ based on two different system identification methods.
We cannot simply compare the goodness-of-fit by comparing the likelihood objectives as they might differ for different identification methods.
Furthermore, the parameter descriptions may also differ (e.g. point estimation vs. set description)
In addition, can we be certain, in a quantitative manner, that the systems behave differently after the propagation through dynamics?
\footnote{We note the Kullback-Leibler (KL) divergence
% , in the field of variational inference and reinforcement learning.
% Mathematically, the KL divergence between two (state) distributions $P$ and $Q$ is
% \[D_{KL} (P \|Q) = -\int \ln \frac{q(x)}{p(x)} dP(x)\approx -\frac1N\sum_{i=1}^N  p(x^i) \ln \frac{q(x^i)}{p(x^i)}.\]
% The right-hand-side is a Monte-Carlo estimate of the KL-divergence given samples \(x_i\sim P, i=1,\dots, N\).
% Though this approach can be related to our discussion via, e.g., the
% integral probability metric (IPM) of \cite{muller1997integral},
% A key difference is that the KL-divergence metric 
is not
\textit{distribution-free}: we need distribution functions to calculate its estimate. 
% In contrast, proposed new concept of metric is \emph{distribution-free}.
}

We summon the strength of KME to endow almost arbitrary data types the meaning of distance through the Hilbert space embedding.
This allows us to compare systems by performing statistical two-sample tests~[cf. \cite{Gretton12:KTT}] using the simulation samples.

Given the state distribution embeddings of two different systems $\mu_{x_t}, \mu_{y_t}$ computed as in \eqref{eq:kme},
we may measure \textit{how different two stochastic systems are} by straightforwardly computing their distance in the embedding Hilbert space,
$\|\mu_{x_t} - \mu_{y_t}\|_{\mathcal H}.$
This quantity is also known as a \textit{maximum mean discrepancy} (MMD). 
Using \textit{kernel trick}~\eqref{eq:trick} and the estimator in \eqref{eq:kme}, we obtain the following. 
\begin{lemma}{(Sample-based estimator for RKHS distance; MMD)}
Given two sets of samples $\{x_i\}_{i=1}^M$ and $\{y_i\}_{i=1}^N$ from simulations of two dynamical systems, a sample-based estimator for $\|\mu_{x_t} - \mu_{y_t}\|_{\mathcal H}$ is given by
  \begin{equation}
	\begin{aligned}
	\| \hat{\mu}_x - &\hat\mu_y\|_{\mathcal{H}}^2  = \frac1{M^2}  \sum_{i,j=1}^M k (x_i,x_j) \\
	- & \frac2{MN}\sum_{i=1}^M\sum_{j=1}^N k (x_i,y_j) + \frac1{N^2}\sum_{i,j=1}^N k (y_i,y_j),
	\end{aligned}
    \label{eq:kernel_trick}
  \end{equation}
\end{lemma}
where we omit time index $t$ for conciseness. More details can be found in \cite{scholkopf2002learning}.

We propose that the goodness-of-fit may again be straightforwardly measured by the RKHS distance 
$\|\mu_{x(\tau,\hat \xi_1)} - \mu_{x(\tau,\hat \xi_2)}\|_{\mathcal H},$
where $x(\tau,\hat \xi_1)$ and $x(\tau,\hat \xi_2)$ are two state distributions under two uncertain  parameter descriptions.

This is powerful in that it can compare arbitrary (unknown) uncertain systems.
We demonstrate this flexibility in Section \ref{sec:exp2}.

\subsection{Uncertainty propagation via KPP}

We have thus far discussed the use of KME as \textit{representation} and \textit{goodness-of-fit measure} for uncertainty in stochastic systems.
In this section, we propose to use KPP for \textit{uncertainty propagation}, which is at the core of many stochastic control algorithms.

We first present two different views of uncertainty propagation in systems. From a statistical standpoint, they correspond to the \textit{diagonal} and \textit{U-statistics} estimation. In particular, the U-statistic estimator is known to have lower variance than the diagonal estimator but to require more samples.
% \footnote{There are also other possibilities in terms of statistics such as a class of incomplete U-statistics which can reduce computational complexity.}
We then show a novel recursive application of the so-called reduced set method in propagating stochastic dynamics forward.
% \jz{ref for diagonal/U/UMVU statistics. @krik: Is there a better way to explain the advantage of using U-stats here? e.g. variance reduction and why? Is this U-stats UMVU? }
% \km{Yes, it is UMVU against a certain class of estimators. U-statistic is the most natural estimator in this setting. It is also unbiased. We can also mention other possibilities such as a class of incomplete U-statistics which can help reduce computational complexity. }

\subsubsection*{Direct propagation via KPP:}
The \textit{main idea} of this algorithm is simple: sample a realization of the uncertainty, and evolve the system as it is deterministic. The steps are presented in Algorithm \ref{alg:prop1}. By doing so, we view the deterministic evolution as algebraic operations performed on the uncertainty.
\begin{algorithm}[tbp]
	\caption{Direct KPP for uncertainty propagation}
	\label{alg:prop1}
	\begin{algorithmic}[1]
		\STATE \textbf{Given}: initial state $x(0)$, (uncertain) dynamical system $f(x, t, \xi)$.
		% \LOOP
		\STATE \textbf{Output}: KME estimate $\hat\mu _ {x(\tau,\xi)}$ at time $\tau$.
		\STATE Choose realization of the uncertain variable nodal set \colnode\ either via collocation or sampling.
		\STATE Evolve the deterministic system forward, either via difference equation or numerical integration, obtain the states $\{x(\tau, \xi^i)\}_{i=1}^N$ at time $\tau$. 
		\label{step:prop}
		\STATE Compute KME estimate $\hat \mu _ {x(\tau,\xi)}$ by \eqref{eq:kme}.
		% \ENDLOOP
	\end{algorithmic}
\end{algorithm}

% talk about cts time
In Step \ref{step:prop},
% propagation is done differently depending on whether the underlying system is discrete or continuous-time. For discrete time, the evolution is rather straightforward as
% \begin{equation}
% \label{eq:dt_sys}
% x^+ = f(x, \xi, t).
% \end{equation}
if the underlying system model is continuous-time, we rely on numerical integration to propagate the samples forward. 
Let us consider the integral of the dynamics function (deterministic or random) over the time period $[0,t],\ x(t, \xi) =  x(0) + \int _0^t f(\tau, \xi) d\tau.$
In practice, this integral is often intractable. Numerical integration is performed to approximate its value,
$
\hat x(t, \xi, h) \approx x(t, \xi),
$
where $h$ may denote the step size of an one-step numerical integration rule.

One immediate question is, how will the integration error affect the embedding estimate?
I.e., is the following true?
\begin{equation}
\hat \mu _ {\hat x(t,\xi, h)} \to \mu _ {x(t,\xi)},\ \forall t.
\end{equation}
By virtue of Lemma~\ref{thm:kme_consis}, we obtain the following result.
\begin{lemma}{(Consistency
\footnote{With a slight overload of terminology, we note the term \textit{consistent} is used in both statistics and numerical analysis community. In both fields, they refer to the asymptotic convergence of statistical estimator and numerical integration respectively.} of KPP estimation with numerical integration)}
Suppose $k$ is a continuous positive-definite kernel, \colnode\ is chosen either via \textit{i.i.d} sampling or collocation rules. The KPP estimator $\hat \mu _ {\hat  x(t,\xi)}$ produced by a one-step numerical integration rule with step size $h$ in Algorithm \ref{alg:prop1} is consistent, i.e.,
\begin{equation}
\hat\mu_{\hat  x(t,\xi, h)} \to \mu_{x(t,\xi)}, \forall t,\text{ as } N\to\infty,\ h\to 0.
\label{eq:kpp_consist}
\end{equation}
\vspace{-0.5cm}
% cvgs rate
% Furthermore, the convergence rate is of order
% \(\mathcal O (h^p) + \mathcal O (\frac1{\sqrt{N}})\), where $p$ is the order of numerical integration rule and $N$ is the number of samples for the estimator.
\label{thm:kpp_consist}
\end{lemma}
The proof (given in the appendix) is a direct consequence of the consistency of numerical integration and that of the KPP estimator.
% wrap up this sub sec on simple mc like method
Similar propagation methods are used in stochastic collocation in conjunction with gPC [\cite{Xiu2009}]. In the above algorithm, the propagated samples are used to represent the distributions via the RKHS embeddings. In the next section, we shall see a non-trivial generalization of the above algorithm.
% \jz{Importance of stability of numerical integration
% For extreme example, let's consider an stiff ODE. 
% This obviously violates the Lipschitz assumption. However, assuming the system is cts (ck), we may get he ...
% This is different from the view presented in KME15, where recursive operation of reduced set is needed to handle the explosion of data size.}

\subsubsection{Recursive reduced set KPP for uncertainty propagation: }
One nuance is encountered when considering more general descriptions of uncertainty other than the simple parameter uncertainty. For simplicity, let us restrict the discussion to discrete-time dynamics and assume that the uncertainty $\xi$ enters as discrete realizations $ \{\xi_t^i\} $ of stochastic processes at each time $t$.
% , i.e. $i$-th sample of uncertainty at time $t$ is given by $ \xi_t^i $. 
In this case, the system state at time $t$ is a function of all previous-step uncertainties, 
$$
X(t) = G(x_0, \xi_1 , \xi_2 , \dots , \xi_{t-1} ).
$$
From a statistical standpoint, this is a transformation of multiple RVs. To estimate the KME, one can either use the \textit{diagonal estimator} which corresponds to the already-discussed Algorithm \ref{alg:prop1},
$$
\mu^{d} _ {X(t)} = \frac1N \sum_{i=1}^N k (G(x_0, \xi_1 ^i, \xi_2 ^i, \dots , \xi_{t-1}^i ), \cdot ),
$$
or the \textit{U-statistics estimator} which delivers smaller variance
\begin{equation}
\mu^U _{X(t)} = \frac{1}{N^{t}} \sum_{i_1=1}^N  \cdots \sum_{i_t=1}^N k (G(x_0, \xi_1 ^{i_1}, \xi_2 ^{i_2}, \dots , \xi_{t}^{i_t} ), \cdot). 
\label{eq:ustats}
\end{equation}
The downside of the U-statistics estimator is that it may involve exponentially many samples of the uncertain random variable (disturbances).
% kpp 15 paper
To relieve this sample complexity while still capturing the statistical distribution, \cite{Scholkopf2015} proposed to use the \textit{reduced set} method to compute multi-step transformations of RV with only a subset of samples. Intuitively, the reduced set method seeks to find a (small) set of expansion points and weights $\{(x^i, \alpha^i)\}_{i=1}^{N_R}$ such that the expansion $\hat \mu^R_x =   \sum_{i=1}^{N_R} \alpha^i k(x^i, \cdot )$ approximates a U-statistics estimation such as in \eqref{eq:ustats}.

We propose the \textit{recursive reduced set kernel probabilistic programming} for uncertainty propagation in Algorithm \ref{alg:reduced}. 
% The output of this algorithm is the reduced-set embedding of the desired state distribution at time $T$,
% \begin{equation}
% \hat \mu^R_{x_{T}} =   \sum_{i=1}^N \alpha_{T}^i k(x_{T}^i, \cdot ).
% \end{equation}
% \jz{@krik: could you verify that $\alpha_i$ doesn't need to sum to 1?}
\begin{algorithm}[tbp]
	\caption{Recursive reduced set KPP for uncertainty propagation}
	\label{alg:reduced}
	\begin{algorithmic}[1]
		\STATE \textbf{Given}: initial state $x(0)$, (uncertain) dynamical system $x^+ = f(x, t, \xi)$, desired size for reduced-set $N_R$\\
		\textbf{Output: } Reduced set expansion for KME at time $T$
		\begin{equation*}
			\hat \mu^R_{x_{T}} =   \sum_{i=1}^N \alpha_{T}^i k(x_{T}^i, \cdot ).
		\end{equation*}
		\LOOP
		\STATE At time $t$, given the reduced set expansion for embedding the current state distribution, sample $N_\xi$ realizations of the uncertain process
		$$
		\xi_{t}^{j} \sim P_{\xi_t}, j=1,\dots, N_\xi.
		$$
		\STATE Compute KME of next state with U-statistics
		\begin{equation}
		\hat \mu_{x_{t+1}} =   \frac1N \sum_{i=1}^{N_R} \sum_{j=1}^{N_\xi} \alpha_t^i k(f(x_t^i, \xi ^j), \cdot).
		\end{equation}
		\STATE Construct the reduced set $\{(x_{t+1}^i, \alpha_{t+1}^i)\}_{i=1}^{N_R}$ for the next time step
		\begin{equation}
		\hat \mu^R_{x_{t+1}} =   \sum_{i=1}^{N_R} \alpha_{t+1}^i k(x_{t+1}^i, \cdot),
		\end{equation}
		by minimizing the following optimization criterion
		\begin{equation}
		\| \hat \mu^R_{x_{t+1}} - \hat \mu_{x_{t+1}}\|_{\mathcal H}.
		\end{equation}
		\label{step:reduce}
		\ENDLOOP
	\end{algorithmic}
\end{algorithm}

Intuitively, at every time step, we look for a subset of all samples (of the U-statistics samples) to serve as new expansion points. One step of our recursion is similar to the basic idea of efficient quadrature rule [\cite{gauss1815methodus}].
% on opt in reduced set
The optimization problem in Step \ref{step:reduce} has two main tasks, finding the expansion points and weights $\{(x_{t+1}^i, \alpha_{t+1}^i)\}_{i=1}^{N}$ simultaneously. 
There is a wide range of techniques for treating this problem (see Chapter 18 of \cite{scholkopf2002learning}) 
In Section~\ref{sec:experiment}, we give a numerical example as a proof of concept for this procedure. 
% We leave a mathematically rigorous analysis of this optimization problem as well as the theoretical guarantees for recursive reduced set algorithm for multiple steps to future work.
% \footnote{\cite{Scholkopf2015,SimonGabriel2016} proved the consistency of the one-step reduced set embedding estimators. \jz{@krik:verify}}

\section{Numerical experiments}\label{sec:experiment}
We present
% numerical examples as a proof-of-concept.
numerical examples that vary in their uncertain system descriptions and types of tasks, showcasing the flexibility of the proposed framework.  

\subsection{Uncertain ODE}
\label{sec:sode}
\begin{figure}
	\centering
	\includegraphics[width=0.75\columnwidth]{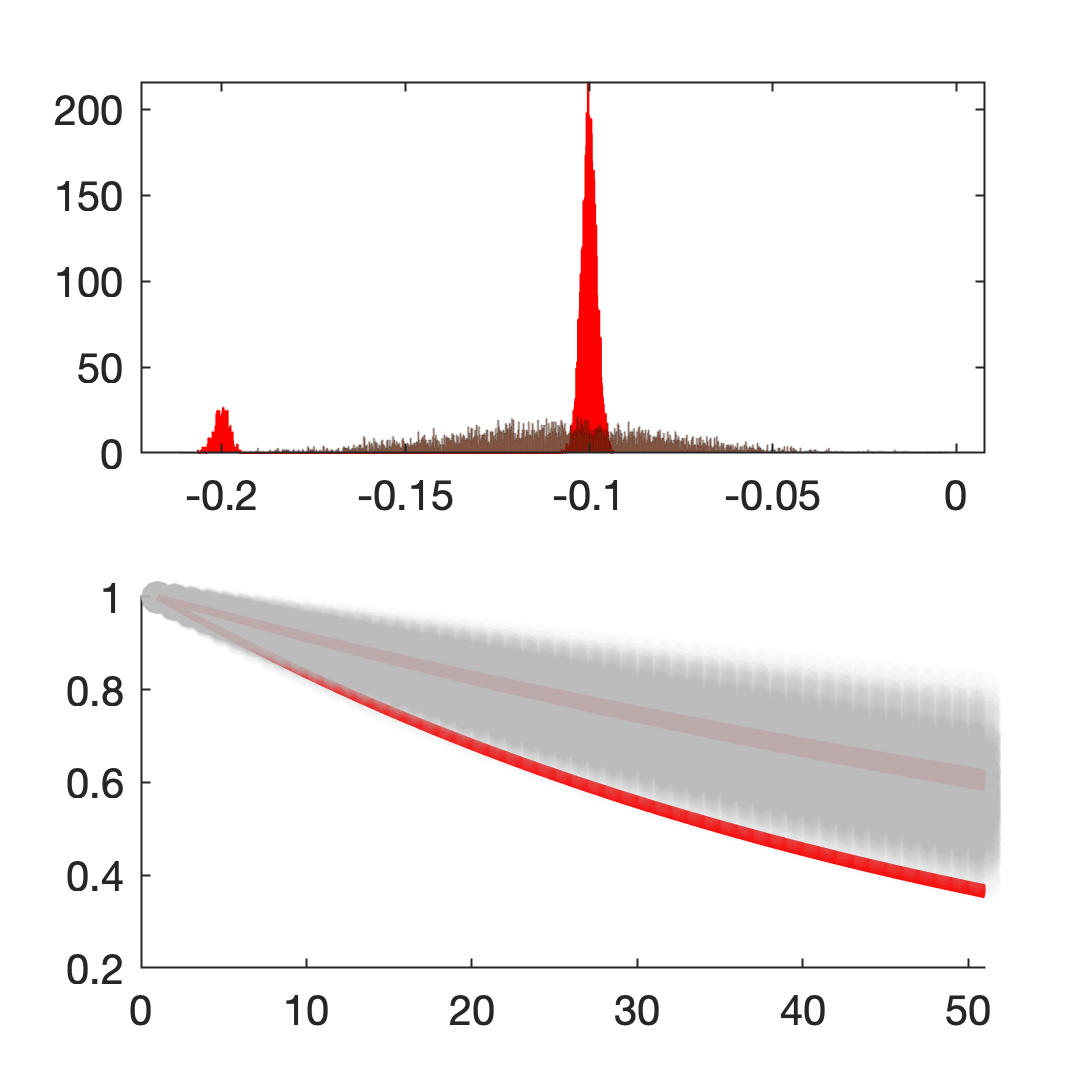}
	% label for 1st fig
	\put(-120,92){\small{parameter value}} % use put from now on
	\put(-190,130){\small{\rotatebox{90}{PDF}}} % use put from now on
	% label for 2nd fig
	\put(-110,0){\small{time}} % use put from now on
	\put(-190,30){{\small\rotatebox{90}{system state}}} % use put from now on

	\caption{(top). Histogram approximating the uncertain parameter density for IVP example. (Red) the histogram represents the initial parameter $\xi \sim \mathrm{GMM}$, (Gray) the histogram represents $\ \xi^\prime \sim N(m, \sigma)$. The two distributions match up to the second moment. However, the GMM is skewed.
	% \jz{consider combine w fig.3 as they are for same experiments}
	(bottom). Two stochastic ODE states with parameter $\xi$ (red) and $\xi '$ (gray). The states are obtained as a result of forward Euler integration.
	}
	\label{fig:init:similar}
\end{figure}
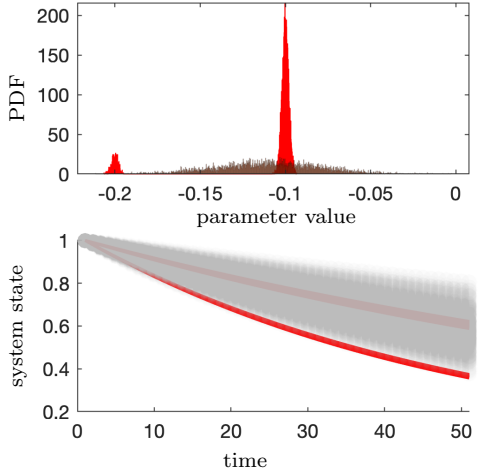

Let us revisit the example of stochastic ODE in \eqref{eq:motivEx},
\begin{equation*}
	% \begin{aligned}
	\dot x(t)  = \xi\ x,\quad x(0) = x_0,
	% \end{aligned}
	% \label{eq:motivEx}
\end{equation*}
% For readers' convenience, we restate the problem formulation.
% $$
% \begin{aligned}
% \dot x(t)  &=& &\xi\cdot\ x(t),&\\
% x(0) &=& &x_0&,
% \end{aligned}
% $$
% where $\xi$ is an RV.
% Different from typical system identification settings, we do not make distributional assumptions about the unknown parameter $\xi$. 
In this experiment, the uncertain \(\xi\) is drawn from a \emph{mixture-of-Gaussian} distribution. 
We then draw another $\xi^\prime$ from a unimodal Gaussian. Its mean and variance are chosen such that the first two moments match those of $\xi$, i.e., 
\begin{equation}
\begin{array}{ll}
\xi \sim \mathrm{GMM}, \ \xi^\prime \sim \mathrm N(m, \sigma)\\
\mathbb E \xi = \mathbb E \xi ^\prime, \ \mathbb E \xi^2 = \mathbb E {\xi^\prime}^2.
\end{array}
\label{eq:gmmmoment}	
\end{equation}
Their distributions are illustrated in Figure \ref{fig:init:similar}~(top).

We wish to answer: \emph{can we quantify the different behaviors of those systems without imposing distribution assumptions?}
To this end, we apply Algorithm \ref{alg:prop1} to obtain two sets of propagated system states $\{x(t, \xi ^i )\} _{i=1}^N$ and $\{x(t, {{\xi'} ^i} )\} _{i=1}^N$.
The state KME estimator $\hat \mu_{X_t}$ and $\hat \mu_{X_t '}$ are computed using \eqref{eq:kme}.

As illustrated in Figure \ref{fig:init:similar}~(bottom), the two moment-matched parameter distributions result in distinct system behaviors over time.
%  This is certainly \textit{not} captured by simply looking at the first two moment statistics of the parameter $\xi$ and $\xi'$ (as they match in \eqref{eq:gmmmoment}).
% \begin{figure}
% 	\centering
% 	\includegraphics[width=0.8\columnwidth]{fig/sys.png}
% 	\put(-110,0){{time}} % use put from now on
% 	\put(-200,50){{\rotatebox{90}{system state}}} % use put from now on
% 	\caption{Two stochastic ODE states with parameter $\xi$ (red) and $\xi '$ (gray). The states are obtained as a result of forward Euler integration.}
% 	\label{fig:sys:similar}
% \end{figure}
To quantitatively compare the behaviors, we compute the RKHS-distance \textit{over time} between those two embeddings $\| \hat \mu_{X_t} - \hat \mu_{X_t '} \|_\mathcal H$ using \eqref{eq:kernel_trick}.

\begin{figure}
	\centering
	% \begin{picture}(10,6)
	\includegraphics[width=0.75\columnwidth]{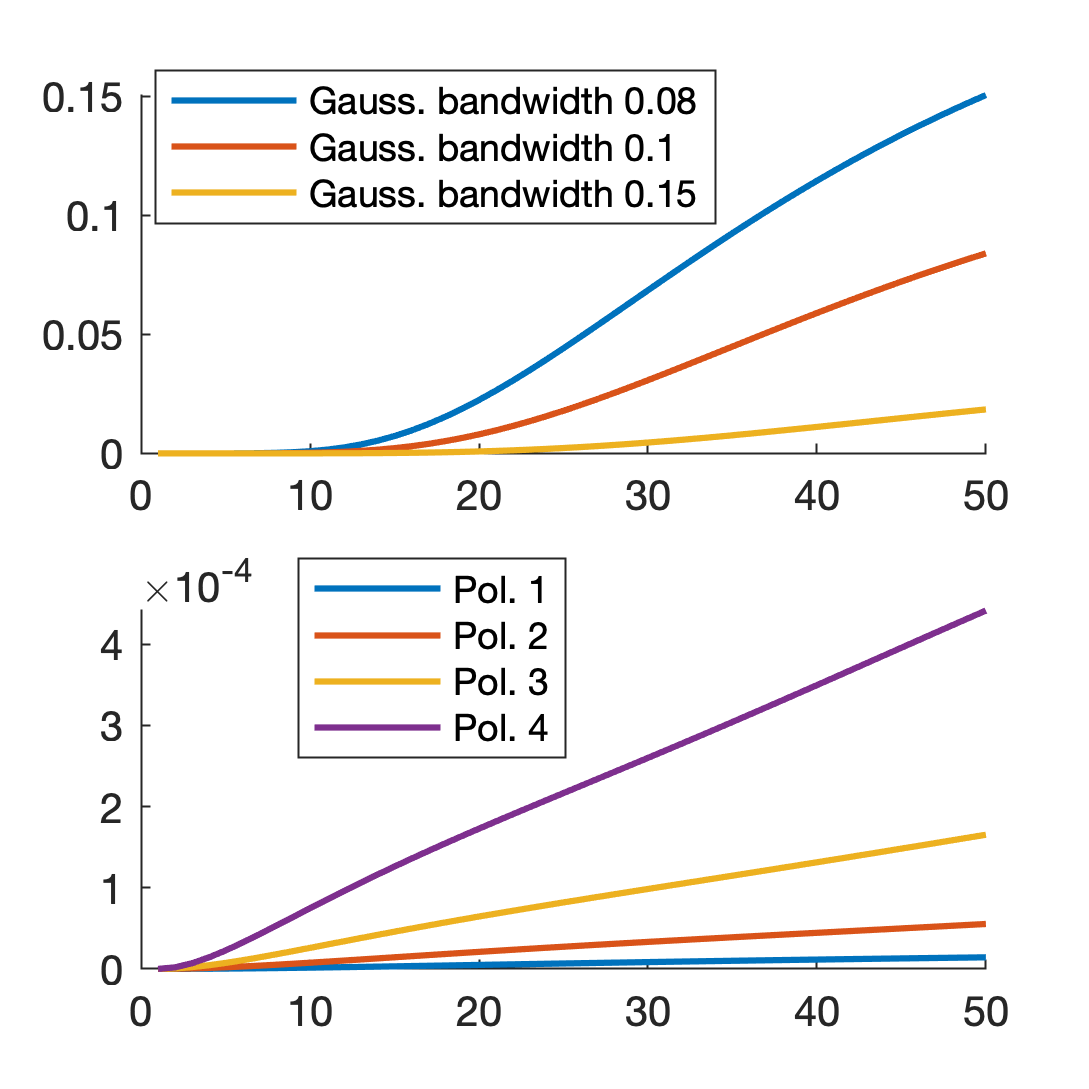}
	\put(-100,0){\small{time}} % use put from now on
	\put(-190,60){\small{\rotatebox{90}{RKHS distance}}} % use put from now on
	\caption{(top): RKHS distance \textit{over time} between two embeddings $\| \hat \mu_{X_t} - \hat \mu_{X_t '} \|_\mathcal H$ associated with Gaussian kernel of different bandwidth.
	(bottom): Embedding distances associated with polynomial kernel of order $1-4$.
	}
	\label{fig:sys:pol}
\end{figure}

To understand how different kernels preserve statistical properties of the system, we also plotted the embedding distances associated with polynomial kernel of order $1-4$ in Figure \ref{fig:sys:pol} (bottom). 
% As discussed, they are equivalent to comparing the first $1-4$ moments using Monte Carlo estimation and then expressing the differences in a unified metric. 
We clearly observe that different polynomial kernels capture different orders of moment information. Notably, the two system states seem to have similar means and therefore the first order polynomial kernel does not differentiate the two systems. As we increase the order of the polynomial kernel, the difference becomes evident.
We then show the Gaussian kernel embeddings in Figure \ref{fig:sys:pol}~(top), where we vary the bandwidth parameter $\sigma$. 
It can be shown that the Gaussian kernel keeps track of statistical moments up to infinite order. 
If bandwidth is large, the kernel treats everything the same so RKHS distance is small.
On the other hand, if bandwidth is small, the kernel treats everything as different. In the limiting case as the bandwidth $\to 0$, it can be shown the KME estimation reduces to the usual Monte Carlo estimation [cf. Section 3.3 in \cite{Scholkopf2015}].
% \begin{figure}
% 	\centering
% 	\includegraphics[width=0.9\columnwidth]{fig/rbf.png}
% 	\put(-120,5){{time}} % use put from now on
% 	\put(-220,70){{\rotatebox{90}{RKHS distance}}} % use put from now on
% 	\caption{RKHS distance \textit{over time} between those two embeddings $\| \hat \mu_{X_t} - \hat \mu_{X_t '} \|_\mathcal H$ associated with Gaussian kernel of different bandwidth.
% 	}
% 	\label{fig:sys:rbf}
% \end{figure}

\subsection{Distribution-free goodness-of-fit for identification}
\label{sec:exp2}
To demonstrate the proposed technique is a good measure of \textit{goodness-of-fit} for arbitrary uncertainty distributions, we apply it to
% a different setting from the previous section. We consider 
% the SDP-based parameter variability estimation 
PVE example in \cite{Mohammadi2015}. 
Different from the uncertainty description in Section~\ref{sec:sode}, this is
typically a ``worst-case'' scenario where the model needs to account for the worst case realization of disturbances, described by ellipsoidal uncertainty sets. 

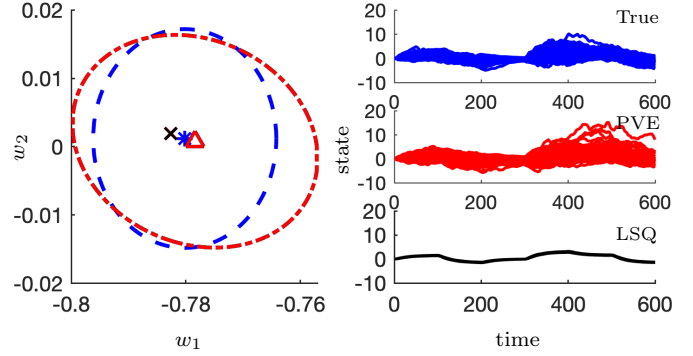
\begin{figure}

	% \begin{tabular}{c}
		\centering
		\includegraphics[width=\columnwidth]{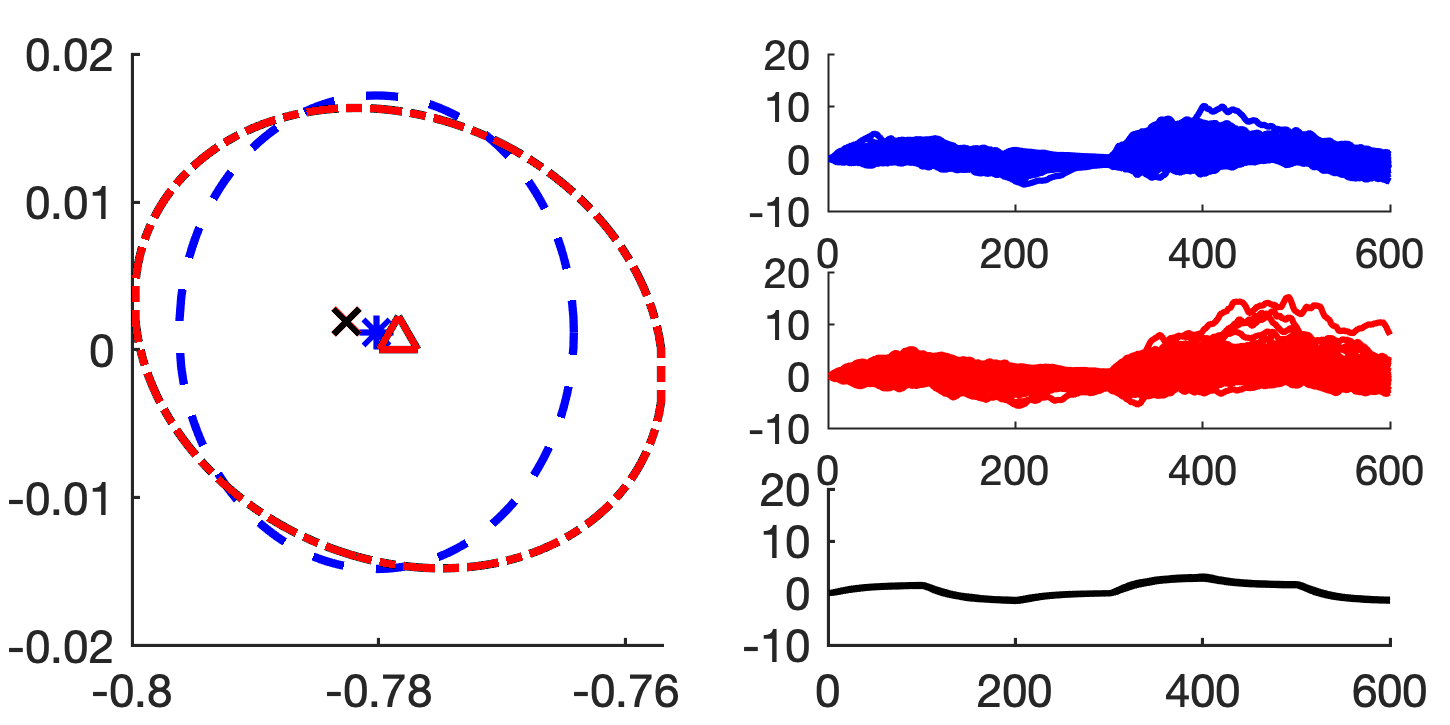}
		% \includegraphics[width=0.5\columnwidth]{fig/ellip_fig_polished/evo.png}
		% mark models
		\put(-25,72){\scriptsize{PVE}}
		\put(-25,112){\scriptsize{True}}
		\put(-25,30){\scriptsize{LSQ}}
		\put(-190,-10){\small{$w_1$}} % use put from now on
		\put(-250,60){\small{\rotatebox{90}{$w_2$}}} % use put from now on	
		% labels for right
		\put(-70,-10){\small{time}} % use put from now on
		\put(-130,55){\small{\rotatebox{90}{state}}} % use put from now on	

	\caption{(left) plots the true uncertainty set (blue), estimated ellipsoidal uncertainty set (red), as well as the LSQ point estimate (black).
	(right) plots the result of evolving the system for 600 steps using true ellipsoid (blue), estimated variability ellipsoid using PVE  (red), Gaussian distributed parameter variability from LSQ (black).
	% % \jz{consider combine with fig below}
	}
	\label{fig:pve_merge}
\end{figure}

In this example, the model to be identified is a time-varying autoregressive exogenous-input model (ARX) with variable parameter. 
\begin{equation*}
y_{k} = a^1y_{k-1} + a^2y_{k-2} + b^1u_{k-1}. 
\label{equ:2Osys}
\end{equation*}
The uncertain time-varying parameters are $w = ( a^2, b^1)$.
To generate the data, we follow the setup of \cite{Mohammadi2015} to choose the true uncertain parameters \(w^{i}\)
uniformly randomly\footnote{We do not use this information during our measuring goodness-of-fit. In addition, we note the original data generation procedure may be extended beyond uniform sampling as robust optimization only concerns the distribution support.} within the true ellipsoidal uncertainty set \(\mathcal E (\bar w , Z)\).
The identification was performed according to that paper.
As a result, we obtain the estimated ellipsoid $\mathcal E ( w_\text{PVE}, Z_\text{PVE})$ and the LSQ parameter $w_\text{LSQ}$
% and the associated standard deviation of underlying Gaussian distribution $\sigma$
, as illustrated in Figure \ref{fig:pve_merge} (left).
%  the estimated ellipsoidal uncertainty set as well as the LSQ point estimate. The state trajectory and the prediction delivered by both estimators are shown in Figure \ref{fig:states}.
% \begin{figure}
% 	\centering
% 	\includegraphics[width=0.8\columnwidth]{fig/state_plot.png}
% 	\put(-110,0){{time}} % use put from now on
% 	\put(-200,70){{\rotatebox{90}{state}}} % use put from now on
% 	\caption{state trajectory and the prediction delivered by both estimators. (blue) true system states. (black) prediction from PVE. (red) prediction from LSQ.}
% 	\label{fig:states}
% \end{figure}

\begin{figure}
	\centering
	\includegraphics[width=\columnwidth]{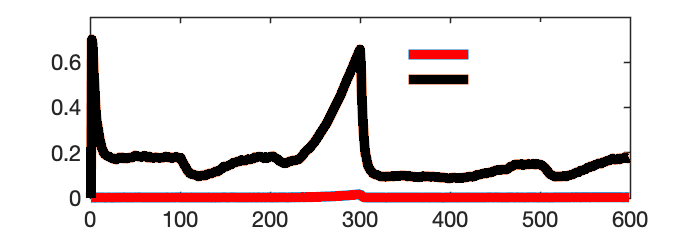}
	% mark the legend
	\put(-81,63){\scriptsize{$\|\mathrm{PVE} - \mathrm{True}\|_\mathcal H$}} 
	\put(-81,53){\scriptsize{$\|\mathrm{LSQ} - \mathrm{True}\|_\mathcal H$}}
	% use put from now on
 	\put(-130,-10){\small{time}} % use put from now on
	\put(-250,15){\small{\rotatebox{90}{RKHS distance}}} % use put from now on
	\caption{We show goodness-of-fit using $\| \hat \mu_{X(t, \hat \xi_\mathrm{PVE} )} - \hat \mu_{X(t, \xi ^*)} \|_\mathcal H$ for estimated parameter variability using PVE (red), and $\| \hat \mu_{X(t, \hat \xi_\mathrm{LSQ} )} - \hat \mu_{X(t, \xi ^*)} \|_\mathcal H$ for Gaussian-distributed parameter from LSQ estimation (black). They correspond to the system evolution over time in Figure \ref{fig:pve_merge} (right).}
	\label{fig:mmd_evo3}
\end{figure}
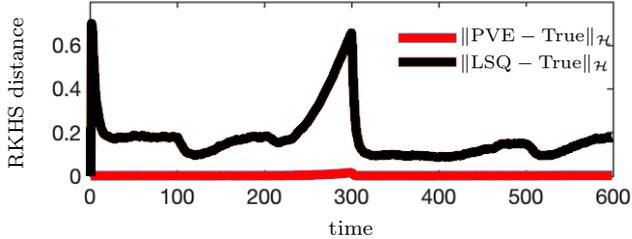

We evolve the system for 600 steps using the true model and those two different uncertain parameter models (PVE vs. Gaussian distributed parameter $w\sim\mathcal N (w_\mathrm{LSQ}, Z_\mathrm{LSQ})$ resulting from the LSQ assumption, where $Z_\mathrm{LSQ}$ is estimated by the sample variance as in that paper). 
The resulting state trajectories are shown in Figure \ref{fig:pve_merge} (right). We apply Algorithm \ref{alg:prop1} and compute the RKHS distance $\| \hat \mu_{X(t, \hat \xi_\mathrm{PVE/LSQ} )} - \hat \mu_{X(t, \xi ^*)} \|_\mathcal H$
between the embeddings of the estimated model and the true model.
% The results are plotted in Figure \ref{fig:mmd_evo3}.
Figure \ref{fig:mmd_evo3} demonstrates that the PVE model (red) matches the true model better than the LSQ model (black) in the sense of RKHS metric. This comparison is performed under no distributional assumptions.

% \subsubsection{The significance of this experiment}
% \begin{itemize}
	% \item
In that paper, it is obvious that LSQ did not deliver the parameter variability that fits the true generating distribution. However, one may ask, is the LSQ estimated parameter $\xi_\mathrm{LSQ}$ nonetheless able to deliver similar system behaviors? 
% After all, Gaussianity assumption often works in practice. 
But there was no unified way to compare them.
%  as they are identified by optimizing \textit{different goodness-of-fit objectives}. 
This paper provides a unifying framework to quantitatively answer this question.

\subsection{Recursive reduced set KPP for uncertainty propagation}
\begin{figure}
	\centering
	\includegraphics[width=\columnwidth]{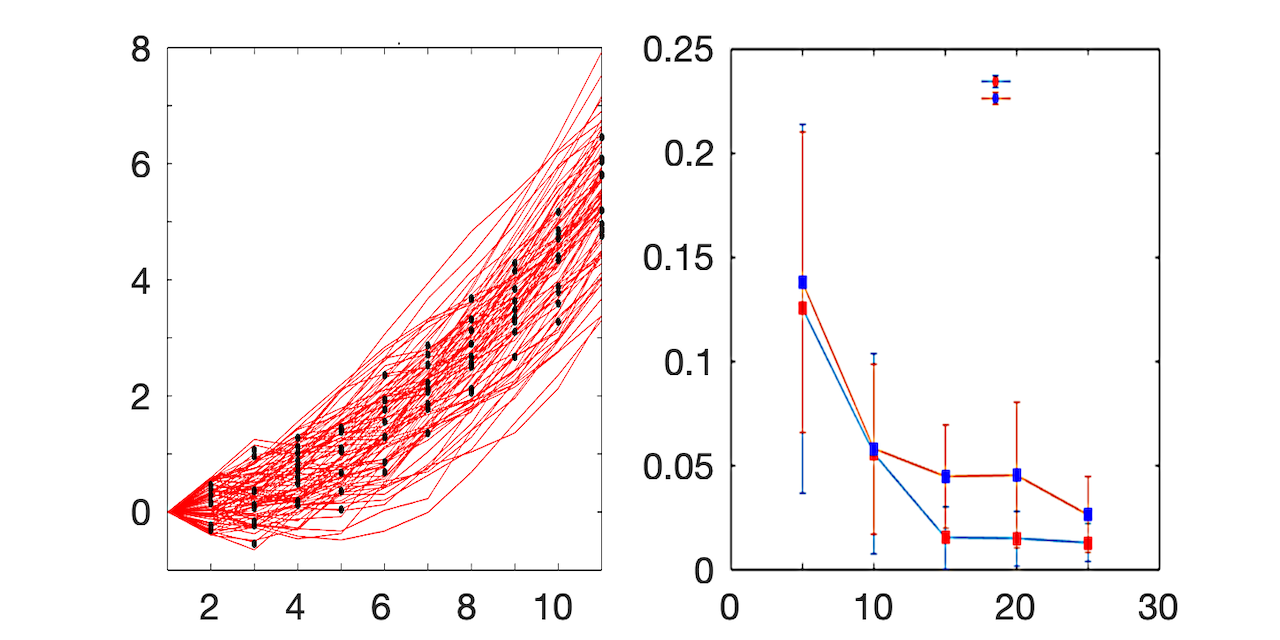}
	\put(-190,-5){\small{time}} % use put from now on
	\put(-235,60){\small{\rotatebox{90}{state}}} % use put from now on
	% rightfig
	\put(-100,-5){\small{number of samples}} % use put from now on
	\put(-130,55){\small{\rotatebox{90}{approx. error}}} % use put from now on
	\put(-50,110){{\tiny alg. 2}}
	\put(-50,105){{\tiny alg. 1}}
	\caption{(left)
	Uncertainty propagation using reduced set method. The uncertainty is forward propagated for 10 steps, using the reduced set methods of size parameter 10. The red lines are Monte Carlo simulations of the system in question. The black dots are chosen reduced set to represent the state distribution at each time step.
	(right)
	Comparing approximation errors of the recursive reduced set algorithm (red, Alg.~\ref{alg:reduced}), $\| \hat \mu_{\tilde x_t} - \mu_{ x_t} \|^2_{\mathcal H}$, with that of the diagonal estimate of KME (blue, Alg.~\ref{alg:prop1})). The dynamics are run for 10 steps. Each method is run 10 times to produce the error bar plot. The state is sampled at time step $t=10$. 
	% This shows that recursive reduced set is able to capture the distribution with fairly small size samples. 
	In calculating the embedding , we use a Gaussian kernel with bandwidth $0.5$.
	}
	% \label{fig:redsetprop}
	\label{fig:merge_red}
\end{figure}
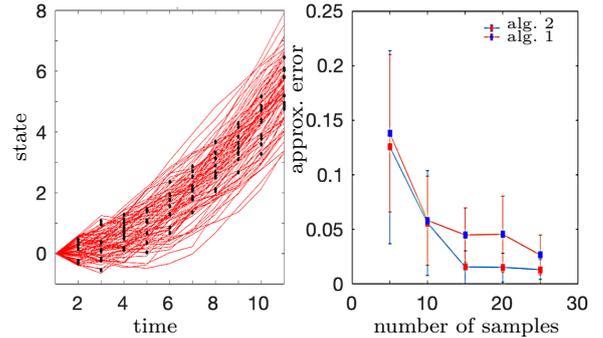

% In the previous sections, we have demonstrated the use of RKHS embeddings as \textit{representation} and \textit{goodness-of-fit} for uncertain dynamical systems.
Relying on the statistical consistency results in Section~\ref{sec:background}, \emph{uncertainty propagation} can be performed straightforwardly using Algorithm \ref{alg:prop1}.
In this section, we demonstrate the use of Algorithm \ref{alg:reduced}.
As a proof of concept, let us consider a simple  discrete-time stochastic system
$$
x(t+1)= x(t) + w(t), w(t) \sim \mathrm{Uniform}\left(-\frac12,\,\,\, \frac12\right) + 0.1  t.
$$
We emphasize the distribution of $w(t)$ could be made fairly arbitrary (and non-stationary)--- the proposed propagation method does not place restrictions on this distribution.
The system evolution is illustrated in Figure~\ref{fig:merge_red}~(left).

At every time step, Algorithm~\ref{alg:reduced} is applied to find an embedding of the current state $\mu_{\tilde x_t} = \sum _{i\in \mathcal R} \alpha _ i k( x_t ^ i, \cdot)$, where $\mathcal R$ denotes the reduced-set indices. Then the dynamics is propagated forward again. Note we use the naive reduced set method following \cite{SimonGabriel2016} which simply samples expansion points $\{x_t^i\}$ from the full set and then solves a quadratic minimization problem for  coefficients $\{\alpha^i\}$ as in Step~\ref{step:reduce}. A more sophisticated method will be introduced in future work. 
As illustrated in Figure \ref{fig:merge_red}~(left), This procedure is repeated for 10 time steps. 
% We then compare the state distribution embedding $\hat \mu_{\tilde x_t}$ from Algorithm \ref{alg:reduced} against that obtained by the diagonal estimate $\hat \mu_{ x_t}$ described in Algorithm \ref{alg:prop1} using the RKHS metric $\|\hat \mu_{\tilde x_t} - \mu_{ x_t} \|^2_{\mathcal H}$.
We compare the RKHS approximation error $\|\hat \mu_{\tilde x_t} - \mu_{ x_t} \|^2_{\mathcal H}$ of the recursive reduced set method in Algorithm \ref{alg:reduced}, against that of the diagonal estimate of Algorithm \ref{alg:prop1}.
While we do not know the true embedding $\mu_{ x_t}$, we approximate it with a large-sample estimate using 500 samples. 
The approximation error in RKHS metric are illustrated in Figure \ref{fig:merge_red}~(right).
We observe a faster convergence in both mean and variance by the recursive reduced set method in Algorithm \ref{alg:reduced}.
% \begin{figure}
% 	\centering
% 	\includegraphics[width=0.8\linewidth]{fig/errbar_red_vs_mc}
% 	\put(-150,0){\small{number of samples used}} % use put from now on
% 	\put(-200,30){\small{\rotatebox{90}{RKHS approx. error}}} % use put from now on
% 	\put(-40,134){{\scriptsize alg. 2}}
% 	\put(-40,127){{\scriptsize alg. 1}}
% 	\caption{Comparing approximation errors of the recursive reduced set algorithm (red, Alg.~\ref{alg:reduced}), $\| \hat \mu_{\tilde x_t} - \mu_{ x_t} \|^2_{\mathcal H}$, with that of the diagonal estimate of KME (blue, Alg.~\ref{alg:prop1})). The dynamics are run for 10 steps. Each method is run 10 times to produce the error bar plot. The state is sampled at time step $t=10$. 
% 	% This shows that recursive reduced set is able to capture the distribution with fairly small size samples. 
% 	In calculating the embedding , we use a Gaussian kernel with bandwidth $0.5$.}
% 	\label{fig:errbarredvsmc}
% \end{figure}

\section{Discussion}
\label{sec:disc}
This paper proposed to use \emph{kernel probabilistic programming} as a unifying tool for treating uncertainties in dynamical systems.
We demonstrated concrete numerical procedures of propagating the uncertainty in dynamics.
% on going work
It is our on-going endeavor to investigate more sophisticated optimization procedures as well as mathematically rigorous analysis of convergence in Algorithm~\ref{alg:reduced} with more general numerical integration. Another important direction is distributionally robust control design and state estimation using RKHS embeddings. 

Compared with existing popular methods such as gPC or GP, RKHS embedding methods for dynamical systems are still in the early development.
This paper serves as a call to action for their wider applications to robust and stochastic control.

\begin{ack}
	We would like to thank Mario Zanon for providing the code to reproduce the PVE experiment.
\end{ack}

\bibliography{ifacconf}             % bib file to produce the bibliography

\begin{thebibliography}{21}
\providecommand{\natexlab}[1]{#1}
\providecommand{\url}[1]{\texttt{#1}}
\providecommand{\urlprefix}{URL }
\expandafter\ifx\csname urlstyle\endcsname\relax
  \providecommand{\doi}[1]{doi:\discretionary{}{}{}#1}\else
  \providecommand{\doi}{doi:\discretionary{}{}{}\begingroup
  \urlstyle{rm}\Url}\fi

\bibitem[{Boots et~al.(2013)Boots, Gordon, and Gretton}]{boots2013hilbert}
Boots, B., Gordon, G., and Gretton, A. (2013).
\newblock Hilbert space embeddings of predictive state representations.
\newblock \emph{arXiv preprint arXiv:1309.6819}.

\bibitem[{Campi et~al.(2019)Campi, Garatti, and Prandini}]{campi2019scenario}
Campi, M.C., Garatti, S., and Prandini, M. (2019).
\newblock Scenario optimization for mpc.
\newblock In \emph{Handbook of Model Predictive Control}, 445--463. Springer.

\bibitem[{Fukumizu et~al.(2004)Fukumizu, Bach, and Jordan}]{Fukumizu04:DRS}
Fukumizu, K., Bach, F., and Jordan, M. (2004).
\newblock Dimensionality reduction for supervised learning with reproducing
  kernel {H}ilbert spaces.
\newblock \emph{Journal of Machine Learning Research}, 5, 73--99.

\bibitem[{Gauss(1815)}]{gauss1815methodus}
Gauss, C.F. (1815).
\newblock \emph{Methodus nova integralium valores per approximationem
  inveniendi}.
\newblock apvd Henricvm Dieterich.

\bibitem[{Gretton et~al.(2012)Gretton, Borgwardt, Rasch, Sch\"{o}lkopf, and
  Smola}]{Gretton12:KTT}
Gretton, A., Borgwardt, K., Rasch, M., Sch\"{o}lkopf, B., and Smola, A. (2012).
\newblock A kernel two-sample test.
\newblock \emph{Journal of Machine Learning Research}, 13, 723--773.

\bibitem[{Gr\"{u}new\"{a}lder et~al.(2012)Gr\"{u}new\"{a}lder, Lever,
  Baldassarre, Pontil, and Gretton}]{Grunewalder12:MDPs}
Gr\"{u}new\"{a}lder, S., Lever, G., Baldassarre, L., Pontil, M., and Gretton,
  A. (2012).
\newblock Modelling transition dynamics in {MDP}s with {RKHS} embeddings.
\newblock In \emph{Proceedings of the 29th International Conference on Machine
  Learning}, 535--542. Omnipress.

\bibitem[{Mohammadi et~al.(2015)Mohammadi, Diehl, and Zanon}]{Mohammadi2015}
Mohammadi, A., Diehl, M., and Zanon, M. (2015).
\newblock {Estimation of uncertain ARX models with ellipsoidal parameter
  variability}.
\newblock \emph{2015 European Control Conference, ECC 2015}, (1), 1766--1771.
\newblock \doi{10.1109/ECC.2015.7330793}.

\bibitem[{Muandet et~al.(2017)Muandet, Fukumizu, Sriperumbudur, and
  Sch\"olkopf}]{Muandet2017}
Muandet, K., Fukumizu, K., Sriperumbudur, B., and Sch\"olkopf, B. (2017).
\newblock Kernel mean embedding of distributions: A review and beyond.
\newblock \emph{Foundations and Trends in Machine Learning}, 10(1-2), 1--141.

\bibitem[{Nishiyama et~al.(2012)Nishiyama, Boularias, Gretton, and
  Fukumizu}]{Nishiyama12:POMDPs}
Nishiyama, Y., Boularias, A., Gretton, A., and Fukumizu, K. (2012).
\newblock {H}ilbert space embeddings of {POMDP}s.
\newblock In \emph{Proceedings of the 28th Conference on Uncertainty in
  Artificial Intelligence}, 644--653.

\bibitem[{Rasmussen and Williams(2006)}]{rasmussen2006gp}
Rasmussen, C.E. and Williams, C.K. (2006).
\newblock Gaussian processes for machine learning.
\newblock \emph{Gaussian Processes for Machine Learning, by CE Rasmussen and
  CKI Williams. ISBN-13 978-0-262-18253-9}.

\bibitem[{Sch{\"{o}}lkopf et~al.(2015)Sch{\"{o}}lkopf, Muandet, Fukumizu,
  Harmeling, and Peters}]{Scholkopf2015}
Sch{\"{o}}lkopf, B., Muandet, K., Fukumizu, K., Harmeling, S., and Peters, J.
  (2015).
\newblock {Computing functions of random variables via reproducing kernel
  Hilbert space representations}.
\newblock \emph{Statistics and Computing}, 25(4), 755--766.
\newblock \doi{10.1007/s11222-015-9558-5}.

\bibitem[{Sch{\"o}lkopf et~al.(2002)Sch{\"o}lkopf, Smola, Bach
  et~al.}]{scholkopf2002learning}
Sch{\"o}lkopf, B., Smola, A.J., Bach, F., et~al. (2002).
\newblock \emph{Learning with kernels: support vector machines, regularization,
  optimization, and beyond}.
\newblock MIT press.

\bibitem[{Scokaert and Mayne(1998)}]{scokaert1998min}
Scokaert, P.O. and Mayne, D.Q. (1998).
\newblock Min-max feedback model predictive control for constrained linear
  systems.
\newblock \emph{IEEE Transactions on Automatic control}, 43(8), 1136--1142.

\bibitem[{Simon-Gabriel et~al.(2016)Simon-Gabriel, {\'{S}}cibior, Tolstikhin,
  and Sch{\"{o}}lkopf}]{SimonGabriel2016}
Simon-Gabriel, C.J., {\'{S}}cibior, A., Tolstikhin, I., and Sch{\"{o}}lkopf, B.
  (2016).
\newblock {Consistent kernel mean estimation for functions of random
  variables}.
\newblock \emph{Advances in Neural Information Processing Systems}, 1(i),
  1740--1748.

\bibitem[{Smola et~al.(2007)Smola, Gretton, Song, and
  Sch\"{o}lkopf}]{Smola07Hilbert}
Smola, A.J., Gretton, A., Song, L., and Sch\"{o}lkopf, B. (2007).
\newblock A {H}ilbert space embedding for distributions.
\newblock In \emph{Proceedings of the 18th International Conference on
  Algorithmic Learning Theory (ALT)}, 13--31. Springer-Verlag.

\bibitem[{Song(2008)}]{Song08:Thesis}
Song, L. (2008).
\newblock \emph{Learning via {H}ilbert Space Embedding of Distributions}.
\newblock Ph.D. thesis, The University of Sydney.

\bibitem[{Song et~al.(2013)Song, Fukumizu, and Gretton}]{Song2013}
Song, L., Fukumizu, K., and Gretton, A. (2013).
\newblock Kernel embeddings of conditional distributions: A unified kernel
  framework for nonparametric inference in graphical models.
\newblock \emph{{IEEE} Signal Processing Magazine}, 30(4), 98--111.

\bibitem[{Sriperumbudur et~al.(2010)Sriperumbudur, Gretton, Fukumizu,
  Sch\"{o}lkopf, and Lanckriet}]{Sriperumbudur10:Metrics}
Sriperumbudur, B., Gretton, A., Fukumizu, K., Sch\"{o}lkopf, B., and Lanckriet,
  G. (2010).
\newblock {H}ilbert space embeddings and metrics on probability measures.
\newblock \emph{Journal of Machine Learning Research}, 99, 1517--1561.

\bibitem[{Tempo et~al.(2012)Tempo, Calafiore, and
  Dabbene}]{tempo2012randomized}
Tempo, R., Calafiore, G., and Dabbene, F. (2012).
\newblock \emph{Randomized algorithms for analysis and control of uncertain
  systems: with applications}.
\newblock Springer Science \& Business Media.

\bibitem[{Xiu(2009)}]{Xiu2009}
Xiu, D. (2009).
\newblock Fast numerical methods for stochastic computations: a review.
\newblock \emph{Communications in computational physics}, 5(2-4), 242--272.

\bibitem[{Xiu and Karniadakis(2002)}]{Xiu2003}
Xiu, D. and Karniadakis, G.E. (2002).
\newblock The wiener--askey polynomial chaos for stochastic differential
  equations.
\newblock \emph{SIAM journal on scientific computing}, 24(2), 619--644.

\end{thebibliography}
% with bibtex (preferred)

%%%%%%%%%%%%%%%%%%%%%% appendix
\appendix
\section{Proof of Lemma \ref{thm:kpp_consist}}  % Each appendix must have a short title.
\begin{pf}
We provide a proof for the consistency of Algorithm \ref{alg:prop1}.
\begin{equation}
\begin{array}{ll}
\| \hat \mu _ {\hat F(x,\xi, t)} - \mu _ {F(x,\xi, t)}\| = \|\hat \mu _ {\hat F(x,\xi, t)} - \hat \mu _ { F(x,\xi, t)} \\
+ \hat \mu _ { F(x,\xi, t)} - \mu _ {F(x,\xi, t)}\|\\
\leq \|\hat \mu _ {\hat F(x,\xi, t)} - \hat \mu _ { F(x,\xi, t)} \| + \| \hat \mu _ { F(x,\xi, t)} - \mu _ {F(x,\xi, t)}\|\\
\leq C\cdot \|{ \hat F(x,\xi, t)} -  {F(x,\xi, t)}\| + \| \hat \mu _ { F(x,\xi, t)} - \mu _ {F(x,\xi, t)}\|\\
\to 0
\end{array}
\label{eq:pf_kpp_consist}
\end{equation}
In the last inequality,
The first term is due to the continuity of kernel $k$.
In the last line, the first term converges to $0$ due to the consistency of numerical integration whereas the second term converges due to the consistency of KPP in Proposition \ref{thm:kpp_consist}.

The convergence rate is
\(\mathcal O (h^p) + \mathcal O (\frac1{\sqrt{N}})\). The first term is
the result of p-th order one-step numerical integration rule (e.g. RK-p) with step size $h$. 
The second term is triggered
by The KPP estimator finite-sample convergence.
$N$ is the sample size used by the KPP algorithm.
\end{pf}

\end{document}